\documentclass[manuscript,screen]{acmart}

\usepackage{times}
\usepackage{soul}
\usepackage{CJKutf8}
\usepackage{url}
\usepackage{amsmath}
\usepackage{subfigure}
\usepackage[utf8]{inputenc}
\usepackage{graphicx}
\usepackage{amsmath}
\usepackage{booktabs}
\usepackage{xcolor}

\usepackage[normalem]{ulem}
\usepackage{algorithm}
\usepackage{algorithmic}
\usepackage{colortbl}
\usepackage[misc]{ifsym}
\usepackage{makecell}
\usepackage{color}
\usepackage{multirow}
\usepackage[breakable]{tcolorbox}
\usepackage{colortbl}
\usepackage{xcolor} 
\usepackage{wrapfig}

\newcommand{\ignore}[1]{}
\usepackage{csquotes}
\definecolor{bleudefrance}{rgb}{0.19, 0.55, 0.91}
\definecolor{yes}{RGB}{239,211,69}
\definecolor{carminered}{rgb}{1.0, 0.0, 0.22}
\definecolor{crimsonglory}{rgb}{0.75, 0.0, 0.2}

\setcopyright{none}
\makeatletter
\def\@copyrightspace{\relax}
\makeatother

\makeatletter
\patchcmd{\@maketitle}{\@copyrightpermission}{}{}{}
\makeatother

\begin{document}

\title{New Paradigm for Evaluating Scholar Summaries: \\A Facet-aware Metric and A Meta-evaluation Benchmark}

\author{Tairan Wang}
\affiliation{
 \institution{King Abdullah University of Science and Technology}
 \country{Saudi Arabia} 
}
\email{tairan.wang@mbzuai.ac.ae}

 \author{Xiuying Chen}
 \authornote{Corresponding authors.}
\affiliation{
 \institution{Mohamed bin Zayed University of Artificial Intelligence}
 \country{UAE} 
}
\email{xiuying.chen@mbzuai.ac.ae}

\author{Qingqing Zhu}
\affiliation{
 \institution{National Institutes of Health}\country{United States} 
}

\author{Taicheng Guo}
\affiliation{
 \institution{University of Notre Dame}\country{United States} 
}

\author{Shen Gao} 
\affiliation{
 \institution{Shandong University}\country{China} 
}

\author{Zhiyong Lu}
\affiliation{
 \institution{National Institutes of Health}\country{United States} 
}

\author{Xin Gao}\authornotemark[1]
\affiliation{
 \institution{King Abdullah University of Science and Technology}\country{Saudi Arabia} 
}
\email{xin.gao@kaust.edu.sa}

\author{Xiangliang Zhang}\authornotemark[1]
\affiliation{
 \institution{University of Notre Dame}\country{United States} 
}
\email{xzhang33@nd.edu}
\renewcommand{\shortauthors}{Xiuying Chen, et al.}

\begin{abstract}
Evaluation of summary quality is particularly crucial within the scientific domain, because it facilitates efficient knowledge dissemination and automated scientific information retrieval. 
This paper presents conceptual and experimental analyses of scientific summarization, highlighting the inadequacies of traditional evaluation methods. These methods, including $n$-gram overlap calculations, embedding comparisons, verification, and QA-based approaches, often fall short in providing explanations, grasping scientific concepts, or identifying key content.
Correspondingly, we introduce the Facet-aware Metric (FM), employing LLMs for advanced semantic matching to evaluate summaries based on different facets.
The \textit{facet granularity} is tailored to the structure of scientific abstracts, offering an integrated evaluation approach that is not fragmented, while also providing fine-grained interpretability.
Recognizing the absence of an evaluation benchmark in the scientific domain, we curate a Scientific abstract summary evaluation Dataset (ScholarSum) with facet-level annotations.
Our findings confirm that FM offers a more logical approach to evaluating scientific summaries.
In addition, fine-tuned smaller models can compete with LLMs in scientific contexts, while LLMs have limitations in learning from in-context information in scientific domains.
We hope our benchmark inspires better evaluation metrics and future enhancements to LLMs: \url{https://github.com/iriscxy/ScholarSum}.
\end{abstract}

%%

\maketitle

\section{Introduction}
\textcolor{black}{The distillation of scientific papers into succinct abstracts is crucial, enabling researchers to quickly grasp the essence of extensive research and promoting the swift exchange of knowledge in an era of information overload~\cite{altmami2022automatic}. 
Scholarly paper summarization involves condensing complex academic content into brief, accessible overviews that capture the core findings, methodologies, and implications of the original research. 
This process not only aids researchers in identifying relevant studies but also facilitates cross-disciplinary understanding, fostering collaborative efforts and stimulating informed discussions that drive scientific discovery forward.
High-quality summarization is particularly vital in domains where rapid advancements demand efficient information retrieval, such as medicine, computer science, and bioinformatics~\cite{hersh2008information}. 
By providing concise yet comprehensive insights, summaries support evidence-based decision-making and accelerate the translation of research into real-world applications. 
Diverse summarization techniques have emerged~\cite{an2021enhancing,zhang2022hegel,koh2022empirical}, leveraging approaches ranging from extractive and abstractive models to hybrid methods. 
The effectiveness of these techniques is increasingly evaluated using multidisciplinary datasets spanning a wide array of fields, such as medicine, engineering, and natural sciences~\cite{cohan2018discourse}, underscoring the importance of domain-specific benchmarks in advancing the state of summarization systems.}

One summarization method is often claimed to perform better than other baselines in terms of evaluation metrics such as ROUGE~\cite{lin2004rouge}, BERTScore~\cite{zhang2019bertscore}, QuestEval~\cite{scialom2021questeval} and ACU~\cite{Liu2022RevisitingTG}. 
However, how can we evaluate whether these metrics truly reflect the comprehensibility and fidelity of the generated summary to the original paper's core content in the scientific domain? 
In this paper, we discuss their limitations and present our evaluation metrics and meta-evaluation benchmark (for the evaluation of evaluation metrics).

\begin{figure*}
\centering
\includegraphics[scale=0.5]{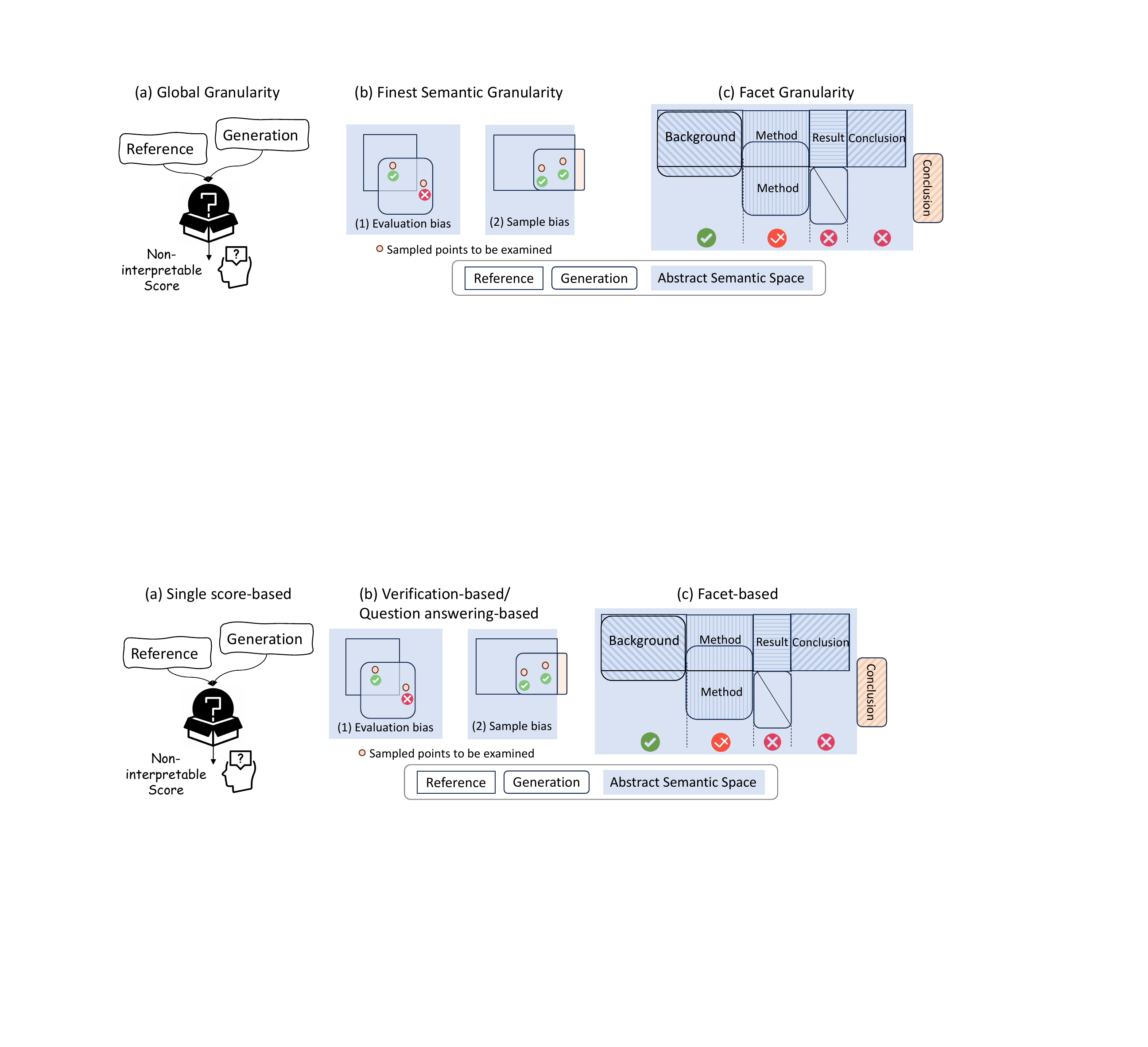}
\caption{
(a) Single score methods like ROUGE and BERTScore lack explainability.
(b) Verification methods like ACUs show \textit{evaluation bias} against correct non-reference claims and \textit{sample bias} in semantic coverage.
(c) FM metric employs LLMs for broader semantic matching and organize evaluation by structured facets. 
The \textit{abstract semantic space} denotes all the semantics that can be found or inferred from the reference.
The yellow square for 'conclusion' indicates the generated conclusion is semantically incorrect. Different textures represent background, method, result, and conclusion sections. The mark after 'Result' in Figure 2 denotes 'partially correct.' 
}
\label{fig:intro}
\end{figure*}

As shown in Figure~\ref{fig:intro}, 
 traditional evaluation methods often fall into two extremes: some of them assess the entire text at a \textit{global granularity}, yielding a single score, for example, ROUGE and BERTScore, which function as black boxes, lacking interpretable reasoning.
Other QA-based and verification methods engage in \textit{finest-grained semantic analysis}, such as QuestEval and Atomic Content Units (ACUs), which compare the generation with reference by sampling and examining the finest semantic units, e.g., the true or false of one fact. 
 This approach inherently limits the evaluation to the sampled content, which may not represent the entire abstract comprehensively. 
 Moreover, this approach can also introduce evaluation bias, as the reference may not fully cover all the correct information, and units sampled outside the reference but within the semantic space do not necessarily indicate inaccuracy.
 Details will be discussed in \S\ref{rethinking}, and these limitations motivate us to design a set of new metrics to capture both the overall quality and the finer details of summarization.

Our proposed Facet-aware Metric (FM) engages a \textit{facet granularity} evaluation tailored for assessing scientific abstract summaries. 
It aims to achieve a balanced assessment that captures both the comprehensive overview and the essential components of the original paper. 
A 50-year survey shows that over 80\% of scientific papers since the 1970s have used the Introduction, Methods, Results, and Discussion structure \cite{sollaci2004introduction}. 
This common structure is used as it allows for clearer understanding, easier replication, and effective evaluation by peers. 
Abstracts should capture these key elements to accurately reflect the original paper's structure and content.
Building on this insight, our FM method, shown in Figure \ref{fig:intro}(c), evaluates the generated summary by comparing it with the original summary regarding Background, Method, Result, and Conclusion (BMRC)\footnote{Our evaluation paradigm is flexible, not only accommodating texts that follow the BMRC structure but also adapting to texts that may be missing certain components, as detailed in \S\ref{paradigm}.}.
The evaluation of each component (facet) is conducted by LLMs, leveraging their broad knowledge and semantic matching abilities. 
The benefits of our paradigm are twofold:
1) It strikes a balance between global and finest granularity, sharing the advantages of an integrated and comprehensive evaluation for each key component of the abstract. 
2) Despite using LLMs, our evaluation method remains cost-efficient. By deconstructing the evaluation into specific aspects, our approach achieves accurate assessments at just 10\% additional cost, much less than the 5x expense of LLM-based metrics like G-Eval~\cite{liu2303g}.

Another contribution of our work is the development of a meta-evaluation benchmark tailored for scholarly summaries.
While there are meta-evaluation benchmarks for general content summaries, such as TAC~\cite{dang2008overview} and RealSum~\cite{bhandari2020re}, scholarly summaries differ significantly from general summaries of social media posts, news articles, and other non-academic content. 
Scholarly summaries must convey the core scientific contributions and the context of the research accurately. 
In addition to quality, fidelity and fact verification, evaluation metrics for scholarly summaries should also cover the thoroughness of research background, the rigor of methodology, the clarity of result presentation, and the depth of the discussion regarding the research's significance and innovation. 
Therefore, a new meta-evaluation benchmark specifically for scholarly summaries is necessary. 
Our ScholarSum comprises thousands of annotations for hundreds of abstracts across various domains from PubMed and arXiv.
The quality of these abstracts, produced by 10 different models, exhibits a wide range of deficiencies. 
Human annotations were meticulously constructed to identify their issues across different aspects.
It provides a stage for measuring the correlation between any automatic evaluation metrics and the human gold-standard. 
Upon a thorough quality analysis of abstracts in ScholarSum using existing metrics, we observed consistent discrepancies between existing automated metrics and human evaluations.
In contrast, our FM metric aligns closely with human evaluations while providing labels at both granular and overall summary levels.
We also encourage researchers to test their metrics in scientific domain on our benchmark.
Lastly, our research uncovers insightful findings shown in Table ~\ref{findings}, highlighting directions for enhancing the scientific summarizaion performance.

Our main contributions can be summarized as follows:\\
$\bullet$  We present a new facet granularity evaluation paradigm that systematically assesses scientific summaries based on key components such as Background, Method, Result, and Conclusion. This approach offers a balanced and interpretable evaluation by combining global and fine-grained insights.

\noindent$\bullet$ We developed ScholarSum, a meta-evaluation benchmark designed specifically for scholarly summaries. It includes thousands of human-annotated abstracts from diverse academic domains, providing a comprehensive dataset for evaluating the alignment between automated metrics and human judgments.

\noindent$\bullet$ Our research highlights key insights into the performance of different LLMs and evaluation metrics. The findings demonstrate that our FM metric closely mirrors human evaluations, offering a more reliable assessment of scholarly summarization compared to traditional metrics like ROUGE and BERTScore.

The remainder of the paper is organized as follows: In Section \ref{sec:related}, we provide a summary of related work. Section \ref{sec:model} introduces the facet-based evaluation paradigm. The construction process of the benchmark dataset using this evaluation method is detailed in Section \ref{benchmark}. We then present the high quality of our benchmark and evaluate different metrics and summarization models based on it in \ref{sec:experiment}. 
In Section \ref{sec:discussion}, we demonstrate that our decomposition method is valuable not only for summary evaluation but also for reading comprehension tasks. 
We provide the dataset release instruction in Section \ref{release}.
Finally, Section \ref{sec:conclusion} concludes the paper and discusses limitations.

\begin{table}[t]
\centering
\caption{Summary of the key findings in our work.}
\begin{tabular}{p{13.2cm}}
\toprule
\S\ref{compare_system} Comparing Summarization Systems:\\
$\bullet$ Larger is not always better: finetuned smaller LLMs rival larger LLMs in scientific contexts. \\
$\bullet$ GPT-3.5 tends to produce text that is easier to understand but often misses critical scientific statistics.
\\ \hline
\S\ref{main_metric} Comparing Evaluation Metrics:\\
$\bullet$ Existing evaluation metrics show a moderate correlation with human scores and a high inter-correlation with ROUGE scores, emphasizing $n$-gram overlap.\\
$\bullet$ The fine-grained semantic analysis metrics are not suitable for complex scholarly summaries.\\
$\bullet$ Our facet-granularity evaluation 
simplifies and excels, moving beyond mere $n$-gram calculation.\\
$\bullet$ LLMs like GPT-4 have limitations in learning from in-context information in the scientific domain.\\
$\bullet$ Facet-grained annotation is beneficial for both evaluating and understanding abstracts.\\

 \bottomrule
\end{tabular} 
\label{findings}
\end{table}

\section{Related Work}
\label{sec:related}
In this section, we summarize the related work on scholarly summarization, automatic evaluation metrics, and meta-evaluation of the metrics, and LLM for summarization.

\subsection{Scholar Paper Summarization}
Automatic summarization for scientific papers has been studied for decades, with earlier research emphasizing document content and favoring extractive methods~\cite{cohan2018scientific, xiao2019extractive}.
\citet{bhatia2012summarizing} propose summarizing figures, tables, and algorithms in scientific publications to augment search results.
While some extractive works~\cite{xu2019scientific,zhao2022hettreesum,chen2022scientific} attempt to leverage document structure, their performance has generally lagged behind that of state-of-the-art abstractive models.
 Recently, abstractive models have shown significant improvements in summarizing scholarly texts.
 For instance, BigBird \cite{zaheer2020big}, employs a sparse attention mechanism that effectively reduces the quadratic dependency to linear for longer sequences for long document summarization. 
 BART \cite{lewis2020bart} is a denoising autoencoder designed for pretraining sequence-to-sequence models. 
 Specifically addressing the challenges of summarizing long documents, BigBird \cite{zaheer2020big}, employs a sparse attention mechanism that effectively reduces the quadratic dependency to linear for longer sequences. 
 Furthermore, LongT5 \cite{guo2022longt5} incorporates attention mechanisms suited for long inputs and integrates pre-training strategies from summarization into the scalable T5 architecture. 
 More recently, LLMs such as Llama \cite{touvron2023llama} have also achieved notable performance in this domain.

\subsection{Automatic Evaluation Metrics}
A growing number of evaluation metrics have also been developed specifically for assessing LLMs across various tasks, including text understanding, question answering, and factuality verification~\cite{liu2024tiny,liu2024vuldetectbench,gao2024shaping,wang2025word}.
Numerous metrics have been proposed to assess summarization models by comparing generated summaries to ground truth references~\cite{sakai2020retrieval,fang2011diagnostic}. 
Earlier metrics, including ROUGE~\cite{lin2004rouge}, METEOR~\cite{banerjee2005meteor}, and SERA~\cite{cohan2016revisiting}, predominantly relied on $n$-gram overlap calculations. 
As pretrained language models advanced, embedding-based metrics such as BERTScore~\cite{zhang2019bertscore} and BARTScore~\cite{yuan2021bartscore} emerged, focusing on vector space representations. Despite their increased sophistication, these metrics often lack intuitive interpretability. Later developments introduced aspect-aware metrics~\cite{scialom2021questeval, kryscinski2020evaluating, fabbri2022qafacteval,tam2022evaluating}, with a predominant focus on faithfulness via question-answering paradigms. Most recently, \citet{liu2023towards} propose a method to extract semantic units from one text and verify them against another, but this approach faces challenges in the scholarly domain due to the inherent difficulty of extracting meaningful semantic units from complex academic texts. In the era of LLMs, G-Eval~\cite{liu2303g} leverages LLMs to evaluate various aspects of summarization, including coherence, conciseness, and grammatical accuracy. Furthermore, \citet{chan2023chateval} introduce a multi-agent LLM-based referee team designed to autonomously discuss and assess the quality of texts. Nevertheless, these broader evaluation frameworks, which emphasize criteria like fluency and coherence, may not fully align with the nuanced demands of professional text evaluation, particularly in areas such as research depth, methodological rigor, result clarity, and the significance of the discussion.

\subsection{Meta-evaluation Benchmarks}
The evaluation methodologies for summarization metrics, or meta-evaluation, depend heavily on well-designed benchmarks. Notable contributions include works like \cite{factualitycheckers} and \cite{gabriel2021go}, which emphasize factuality in generated summaries, particularly within news corpora. \citet{chen2023follow} introduce three benchmark datasets focused on timeline summarization across different domains. 
Similarly, \citet{chern2024large} explore the use of large language models as debating agents to evaluate various tasks such as question answering and code writing. 
\citet{guo2024appls} specifically address plain text summarization, requiring the inclusion of background explanations and the removal of specialized terminology.
In related fields, \citet{liu2021meta} examine three critical aspects of evaluation metrics: reliability, the ability to detect actual performance differences; fidelity, alignment with user preferences; and intuitiveness, capturing properties like adequacy, informativeness, and fluency in conversational search.
Recently, researchers have sought more nuanced assessments. For example, \citet{bhandari2020re} and \citet{Liu2022RevisitingTG} annotate summaries using semantic content units, drawing from the LitePyramid protocols \cite{shapira2019crowdsourcing}. 
However, to our knowledge, these methodologies have yet to be applied to scientific corpora, leaving the effectiveness of these evaluation metrics in scientific domains largely unexplored.
Compared to strategies designed for meta-evaluation, such as those in \cite{chern2024large} and \cite{guo2024appls}, the development of benchmarks for scientific summarization is still in its early stages. Our experiments reveal that evaluating scholarly content is particularly challenging due to its dense information and more complex grammar, which distinguishes it from general-domain texts.

\subsection{LLM for summarization}
The rise of large language models, such as GPT-3~\cite{floridi2020gpt} and GPT-4~\cite{achiam2023gpt}, has led to significant improvements in the quality of generated summaries, including in the academic domain. 
These models are able to process and generate fluent and coherent summaries based on large-scale pretrained knowledge. 
To name a few, \citet{bahrainian2021cats} introduce new mechanism to control the underlying latent topic distribution of the produced summaries, and \citet{cagliero2019elsa} propose a new summarization approach that exploits frequent itemsets to describe all of the latent concepts covered by the documents under analysis and LSA to reduce the potentially redundant set of itemsets to a compact set of uncorrelated concepts.
However, challenges such as hallucination and fidelity to the source content remain. 
For example, \citet{rane2023contribution,tang2024prioritizing} conduct a thorough examination of vulnerabilities in LLM-based agents within scientific domains, shedding light on potential risks associated with their misuse and emphasizing the need for safety measures. 
Recent studies, such as \cite{fonseca2024can, parmar2024towards}, have explored how large language models can be refined for scientific summarization. 
Additionally, \citet{ren2018sentence} study the use of sentence relations, such as contextual sentence relations, title sentence relations, and query sentence relations, to improve the performance of sentence regression.
However, limitations in model accuracy and interpretability continue to hinder their widespread adoption.
Our work explores these models' performance in the context of scientific paper summarization, contributing valuable findings to this growing body of research.

\section{Facet-based Evaluation Metric}
\label{sec:model}

\textcolor{black}{In this section, we first present our rethinking of the weaknesses in existing metrics and then introduce our proposed facet-based evaluation paradigm.}

\subsection{Rethinking on Existing Metrics}

\label{rethinking}
\textit{\textbf{Single-Score Metric.}}
Traditional metrics, ROUGE and BERTScore, are central in evaluating NLP-generated texts. 
ROUGE emphasizes recall by analyzing $n$-gram overlaps between reference and generated summaries, while BERTScore transforms the two texts into vector representations and calculates similarity using pre-trained language models like RoBERTa~\cite{Liu2019RoBERTaAR}.
Both these metrics yield only single scores without providing details on attributes, such as where and why errors occur, thereby reducing their trustworthiness and reliability.
Hence, as shown in Figure \ref{fig:intro}(a), we classify them into global granularity.

\begin{figure*}
\centering
\includegraphics[scale=0.48]{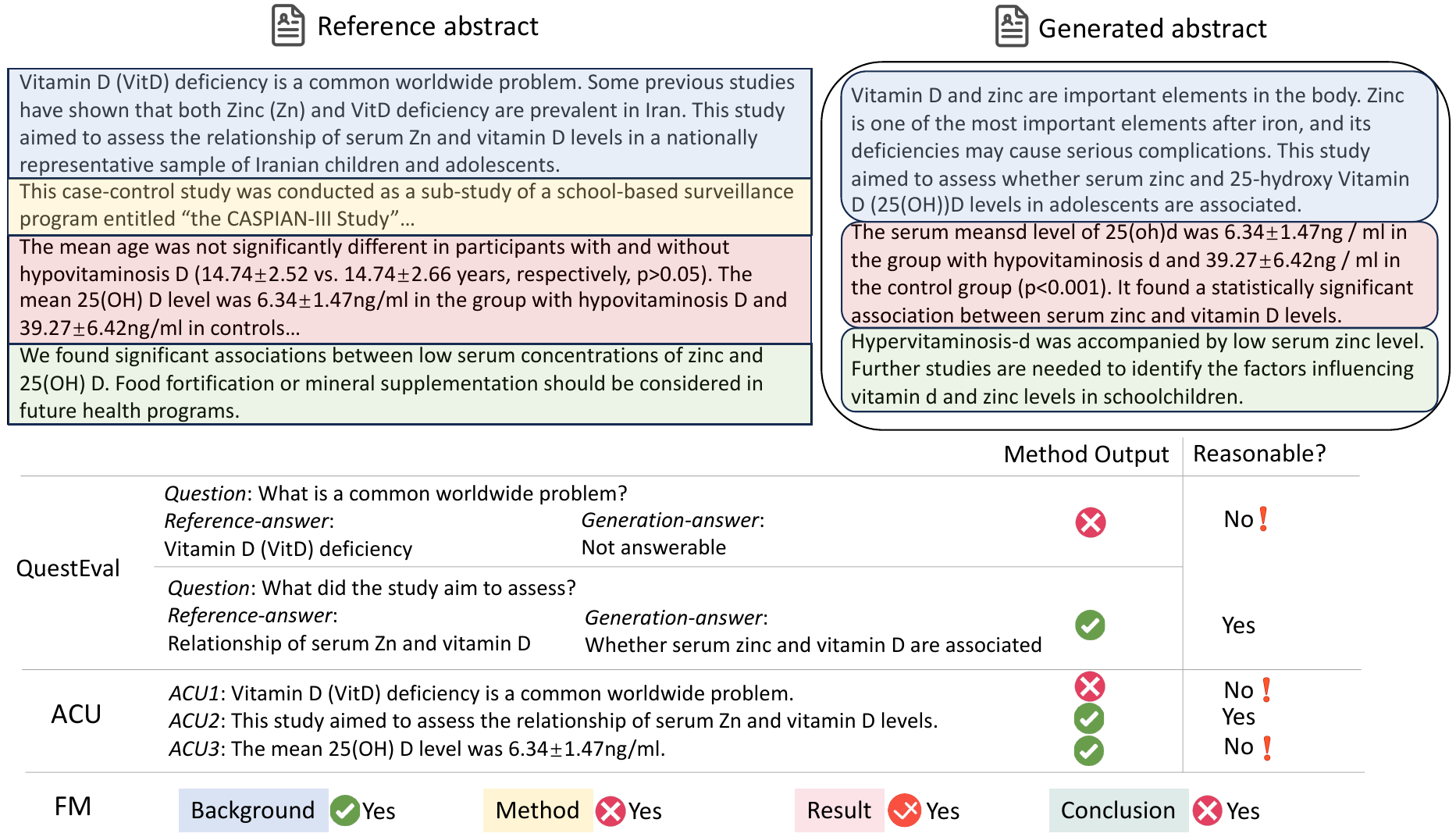}
\caption{
An evaluation case study on PubMed dataset.
QuestEval and ACU require precise alignment between the reference and generated text, which is often unachievable in real-world scenarios.
The semantic segment under scrutiny may not cover the entire semantics of the text either. 
Conversely, our FM metric enables holistic semantic evaluation without segmentation and eases the process by semantic matching across different facets.
}
\label{fig:case}
\end{figure*}

\noindent \textit{\textbf{Question-answering based Metric.}}
Recent explainable evaluation metrics, like QuestEval~\cite{scialom2021questeval}, use question-answering paradigms to assess the accuracy of generated summaries. 
These methods focus on the finest semantic units, such as entities and nouns, as answers, formulating questions to compare answers derived from both the reference and the generated summary.
While this method has a relatively higher correlation with human evaluations, we summarize its limitations as follows.

Firstly, relying solely on a limited set of QA pairs for evaluation risks \textit{sample bias}, as illustrated in Figure \ref{fig:intro}(b).
This is because the meaning of a text is not merely a compilation of isolated facts or data points. 
Rather, it is an interconnected continuum, where ideas, concepts, and nuances interweave.
A limited number of samples might not fully capture this continuum of meaning.
Moreover, discrepancies in answers based on the reference and generated summaries do not necessarily indicate that the generated summary is incorrect, which we name as \textit{evaluation bias}.
For example, as illustrated in the first question in Figure \ref{fig:case}, the generated content fails to directly answer the question and highlight that "Vitamin D deficiency is a common worldwide problem".
However, the generated content still notes the importance of Vitamin D, thereby providing a relevant and informative background introduction.
Nevertheless, according to the QA rule, this would be erroneously labeled as unfaithful content. 
This issue highlights the method's potential drawbacks in accurately assessing summary quality using the question-answering paradigm.

\noindent \textit{\textbf{Claim Verification based Metric.}}
In a related line of work, summaries are evaluated by verifying their claims.
For instance, \citet{kryscinski2020evaluating} extract specific spans from the reference text to assess their consistency with the generated content. 
Building on this, \citet{liu2023towards} introduce ACU, a method that replaces spans with atomic content units, as they find that using smaller annotation units improves annotation consistency.
This technique addresses the previous challenge of formulating appropriate questions for examination.

Yet, the two previous biases still exist.
First, comprehensively covering all points in a sentence, especially in scientific texts characterized by complex structures and specialized terminology, is challenging. 
For instance, consider the reference in Figure \ref{fig:case}, which includes intricate concepts such as "VitD", "Zinc", "25-hydroxy", and "cardiometabolic", along with their relationships. 
Such sentences present a more complicated scenario for segmentation, particularly when compared to the simpler and more straightforward structures in general domains like news.
Secondly, various methods exist to convey similar meanings; however, examining at the unit level necessitates precise alignment, which is not always feasible in practical scenarios.
Thirdly, the accuracy of a unit is context-dependent. 
For instance, consider the ACU3 in Figure \ref{fig:case}. 
Without knowing the group being examined, it's uncertain whether the claim holds true.

We also tried the recent TIGERSCORE~\cite{jiang2023tigerscore}, which is designed to generate evaluation scores with explanations in a sequence-to-sequence approach. 
Unfortunately, this model struggles to produce fluent sentences and fails to yield scores on scholar corpus.
This limitation likely stems from its heavy reliance on its training corpus, which does not include scholarly texts.

\noindent \textit{\textbf{Domain Specific Metrics.}}
To meet the specialized needs of medical reviews, distinct strategies have been proposed. 
For example, \citet{huang2006evaluation} propose a PIO framework, which classifies systematic reviews based on three aspects: \emph{\underline{P}opulation}, \emph{\underline{I}ntervention}, and \emph{\underline{O}utcome}.
Based on this concept, Delta was presented by \citet{wallace2021generating} and later refined by \citet{deyoung2021msˆ2}. 
This method calculates the probability distributions of evidence direction for all I\&O pairs in both the target and generated summaries.
The final score is derived by applying the Jensen-Shannon Divergence to compare these distributions for each I\&O pair, where a lower score indicates a closer alignment with the target summary.
However, abstracts of scientific papers can span various domains, and the PIO components are exclusive to medical reviews.

\subsection{Facet-based Evaluation Paradigm}
\label{paradigm}
Given the previous challenges, we propose a facet-aware evaluation paradigm tailored to the unique attributes of scientific abstracts.
Building on the foundational work~\cite{sollaci2004introduction,dernoncourt2017pubmed,jin2018hierarchical}, which found that the \emph{Background, Method, Result}, and \emph{Conclusion} structure became prevalent since the 1970s, we classify abstract content into distinct facets, forming the BMRC set.
Specifically, `\underline{B}ackground' covers the work's introduction and objectives, `\underline{M}ethod' outlines the experimental methods and comparisons, `\underline{R}esult' discusses observations and data analysis, and `\underline{C}onclusion' includes the conclusions, limitations, and future perspectives.
We reviewed papers on PubMed and arXiv, noting that most abstracts follow the BMRC structure. When they don't, we adjust our evaluation by omitting the non-conforming aspect.

For a quantitative assessment of the alignment between the reference (Input1) and generated abstracts (Input2), they are compared on the \emph{Background} and \emph{Conclusion} facets based on the following prompts:
\begin{tcolorbox}[colback=gray!10, left=1mm, right=1mm, top=1mm, bottom=1mm] 
 - 3: Input2 is generally consistent with Input1.\\
 - 2: Input1 is not mentioned in Input2.\\
 - 1: Input2 contradicts Input1, or Input2 lacks relevant content in this aspect. 
\end{tcolorbox}

Take the \textit{background} section shown in Figure \ref{fig:case} as an example. 
Both the reference and the generated text correctly emphasize the significance of Vitamin D and Zinc (Zn) in the human body, capturing the key elements needed to convey the primary message. 
As a result, the generated summary for the background is assigned a score of 3, reflecting a satisfactory representation of the core information.
In contrast, the generated \textit{conclusion} contains a critical error by incorrectly stating that "hypervitaminosis D is accompanied by low zinc levels," when it should have referred to "hypovitaminosis" instead. 
This mistake misrepresents an essential finding of the study and significantly alters the intended meaning, leading to a much lower score of 1 for this section.

\textcolor{black}{The rating prompt for evaluating }
\emph{Method}/\emph{Result} is:
\begin{tcolorbox}[colback=gray!10, left=1mm, right=1mm, top=1mm, bottom=1mm]
- 4: Input2 generally includes Input1's information, or omits minor details from Input1.\\
- 3: Input2 generally includes Input1's information, but omits a part of the key information from Input1.\\
- 2: Input2 is not empty, but it does not mention any key information in Input1.\\
- 1: Input2 contradicts Input1, or Input2 lacks relevant content in this aspect.
\end{tcolorbox}

Here, 'key information' refers to the essential elements that are critical for understanding the main message of Input1, while 'minor details' consist of less significant supplementary elements whose omission does not drastically impact the overall comprehension. 
For instance, consider the case in Figure \ref{fig:case}. In this example, the generated \textit{result} section points to a correlation between Zinc (Zn) and Vitamin D but fails to specify whether this relationship is positive or negative. This omission results in a score of 3, as knowing the direction of the relationship (positive) is crucial for a complete understanding of the study's conclusions. 
We employ a 3-point scale for general \textit{background} and \textit{conclusion} sections, while a more granular 4-point scale is used for the detailed \textit{method} and \textit{results} sections to capture finer nuances. We recognize that different abstracts may emphasize various aspects of key information. For the purpose of our evaluation, we treat the abstract of the paper as the ground truth, as seen in previous evaluation works~\cite{Liu2022RevisitingTG}. Expanding to multi-reference evaluation remains a subject for future research.

\textcolor{black}{Based on the rating score of each aspect, the overall score of a generated abstract is computed as follows, where the weight of each aspect reflects its relative importance in the scoring process and is introduced in detail in \S\ref{annotation}:
\begin{align} s=(\textstyle \sum^4_{i=1} \text{score}_i/\text{scale}_i \times \text{weight}_i)/4. \label{weight} \end{align} 
For clarity, the weights used here—determined through linear fitting based on the annotation data—are [0.1, 0.3, 0.3, 0.3], with a mean squared error of 0.005, indicating a strong fit. These weights assign different levels of significance to individual components, reflecting their contribution to the overall score.}

\textcolor{black}{We construct a human annotation benchmark dataset following this paradigm, and the details will be introduced in \S\ref{benchmark}.}

\subsection{\textcolor{black}{LLM-based Facet-aware Evaluation}}
\label{llm-based}
Given the proficiency of LLMs in text comprehension, we can utilize them to automatically assign facet-aware scores following our facet-based evaluation paradigm.
Due to the intricacies of comprehending scholarly corpora and to simplify the task, we divide the assessment into two sub-tasks: first, LLM extracts facet-aware segments from both reference and generated abstracts. 
Next, the segments are compared using LLMs, guided by prompts in \S\ref{paradigm}, and a weighted sum is applied to calculate the final score as in Eq.~\ref{weight}.
In our facet-aware metric, we employ GPT-4 to initially extract information of different aspects within the abstract.
The prompt we use is:
\begin{tcolorbox}[colback=gray!10, left=1mm, right=1mm, top=1mm, bottom=1mm] 
What is the background/method/result/conclusion of this work? \\Extract the segment of the input as the answer. \\
Return the answer in JSON format, where the key is background/method/result/conclusion. \\
If any category is not represented in the input, its value should be left empty.
\end{tcolorbox}

\textcolor{black}{Compared to prior evaluation metrics, our evaluation paradigm offers transparency and insight by explicitly extracting and scoring facet-aware segments (e.g., background, method, result, conclusion) independently. }
These segment scores are then combined using a weighted sum to produce the final score. This step-by-step process ensures that readers and users can clearly understand which aspects contributed to the overall score and identify any discrepancies between corresponding facets.

We utilize GPT-3.5, GPT-4, and Llama2 as the foundational LLM for our tasks, denoted as FM (backbone\_name).
We also include variants like FM (backbone w/ few), which adds few-shot examples to measure in-context learning.
The evaluation prompt for background and conclusion facets with in-context samples is as:
\begin{tcolorbox}[colback=gray!10, left=1mm, right=1mm, top=1mm, bottom=1mm,breakable] 
Using a less strict criterion, assess the alignment (1-3) between the two inputs. 

- 3: Input2 is generally consistent with Input1.\\
- 2: Input1 is not mentioned in Input2.\\
- 1: Input2 contradicts Input1.\\
Only return the number. \\

Example 1:\\
Input1: The use of 2-[18f]fluoro-2-deoxy-d-glucose ([18f]FDG) may help to establish the antitumor activity of enzastaurin, a novel protein kinase C-beta II (PKC-II) inhibitor, in mouse xenografts.\\
Input2: Imaging techniques, such as positron emission tomography (PET), are important for diagnosing and monitoring cancer patients. The glucose analogue 2-[F]fluoro-2-deoxy-D-glucose (FDG) is commonly used as a tracer in PET imaging to assess tissue glucose utilization. FDG PET is widely used in diagnosing various types of cancer, and it is being evaluated as a tool to assess the effects of anticancer drugs. Enzastaurin is a novel compound that inhibits protein kinase C-beta (PKC-II), which has been implicated in tumor growth.\\
Number: 3\\

Example 2:\\
Input1: Nissen fundoplication is an effective treatment for gastroesophageal reflux in infants. Laparoscopic procedures after previous laparotomy are technically more challenging. The role of laparoscopic Nissen fundoplication after neonatal laparotomy for diseases unrelated to reflux is poorly described.

Input2: The article discusses the complex nature of gastroesophageal reflux in neonates and infants, which is often caused by a combination of developmental and anatomical factors.\\
Number: 2\\

Example 3:\\
Input1: [18f]FDG PET imaging technique does not correlate with standard caliper assessments in xenografts to assess the antitumor activity of enzastaurin.\\
Input2: These findings suggest that [18F]FDG PET imaging is a useful tool for assessing the antitumor effects of novel compounds, such as enzastaurin, in preclinical studies.\\
Number: 1

\end{tcolorbox}

The evaluation prompt for method and result facets with in-context samples is as:
\begin{tcolorbox}[colback=gray!10, left=1mm, right=1mm, top=1mm, bottom=1mm,breakable] 
% \small
Assess the alignment (1-4) between the two inputs. \\\\- 4: Input2 generally covers the information present in Input1, or omits minor details from Input1.\\- 3: Input2 omits important information from Input1.\\- 2: Input1 is not mentioned in Input2.\\- 1: Input2 contradicts Input1.\\
Only return the number. \\ \\Example 1:\\Input1: We analyzed the methylation status of protocadherin8 in 162 prostate cancer tissues and 47 benign prostatic hyperplasia tissues using methylation-specific PCR (MSP). The patients with prostate cancer were followed up for 15-60 months, and biochemical recurrence was defined as the period between radical prostatectomy and the measurement of 2 successive values of serum PSA level 0.2 ng/ml.\\Input2: The promoter methylation status of protocadherin8 in 162 prostate cancer tissues and 47 normal prostate tissues was examined using methylation-specific pcr (msp). subsequently, the relationships between protocadherin8 methylation and clinicopathological features of prostate cancer patients and biochemical recurrence-free survival of patients were analyzed.\\Number: 4\\
\\Example 2:\\Input1: The present study included 515 patients admitted to the coronary care units or equivalent cardiology wards of the participating hospitals between 2011 and 2012 in north punjab, pakistan. the analysis was focused on identifying the socioeconomic status, lifestyle, family history of mi, and risk factors (i.e. hypertension, diabetes, smoking, and hyperlipidemia). a structured questionnaire was designed to collect data. the lipid profile was recorded from the investigation chart of every patient. for statistical analysis, the kruskal wallis, mann-whitney u, wilcoxon, and chi-square tests were used.\\
Input2: A population-based cross-sectional study was conducted in six regions in north punjab (urban and rural patients). data were collected using trained staff from the patients admitted in coronary care units or equivalent cardiology hospitals in the participating hospitals.\\Number: 3\\\\
Example 3:\\Input1: Hyperglycemia, commencing on the first dose of the steroid given, persisted even after the discontinuation of steroids and improvement of other signs. there were no signs of pancreatitis or type 1 diabetes clinically in laboratory tests. her blood glucose levels were regulated at first with insulin and later with metformin. within 1 year of follow-up, still regulated with oral antidiabetics, she has been diagnosed with type 2 diabetes. \\Input2: The patient was treated with discontinuation of carbamazepine, antihistaminic and systemic steroids, and her hyperglycemia resolved with metformin treatment. The patient's lung, skin, liver, and renal findings regressed, and a patch test with carbamazepine was positive.
\\
Number: 2\\ \\
Example 4:\\
Input1: Hyperglycemia, commencing on the first dose of the steroid given, persisted even after the discontinuation of steroids and improvement of other signs. there were no signs of pancreatitis or type 1 diabetes clinically in laboratory tests. within 1 year of follow-up, still regulated with oral antidiabetics, she has been diagnosed with type 2 diabetes.\\
Input2: The patient recovered without any sequelae.\\
Number: 1

\end{tcolorbox}

We also include the vanilla backbone, which feeds the LLM directly with rating instructions without decomposition.

\section{ScholarSum: A Meta-Evaluation Benchmark for Scholarly Summaries}
%Facet-based Evaluation Benchmark}
\label{benchmark}

To the best of our knowledge, there is currently no meta-evaluation benchmark for the scientific paper summarization domain. Our ScholarSum dataset can be used to assess automatic evaluation metrics, measuring the correlation between various automatic metrics and the human gold-standard.

\subsection{Summarization Systems}

\label{dataset}
\begin{figure}[htb]
    \centering
    \includegraphics[width=1\linewidth]{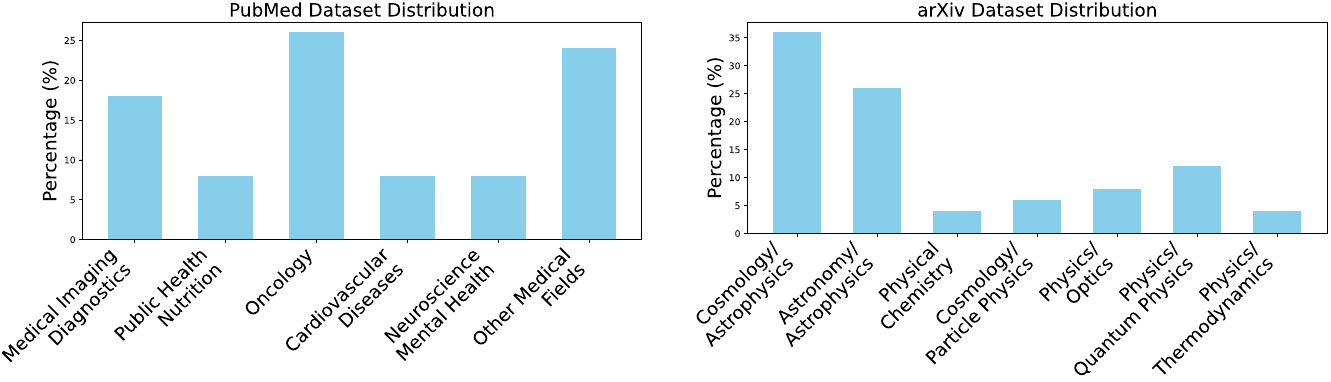}
    \caption{\textcolor{black}{Domain distribution of the PubMed and arXiv datasets.}}
    \label{fig:distribution}
\end{figure}

Our benchmark is constructed using a combination of articles from arXiv and PubMed. ArXiv primarily hosts papers in disciplines such as physics, mathematics, and computer science, whereas PubMed focuses on literature from the biomedical field. To create a diverse evaluation set, we randomly sampled 50 cases from arXiv, ensuring representation across its major categories. For PubMed, we leverage the test set created by~\citet{krishna2023longeval}, which similarly contains 50 cases. This approach allows for a comprehensive evaluation across both scientific and biomedical domains, facilitating a robust comparison between the two sources.
We show the domain distribution of the two datasets in Figure~\ref{fig:distribution}.

For each paper in the arXiv dataset, we employ the pretrained summarization model BART-large~\cite{lewis2020bart}, as well as the recent state-of-the-art model Factsum~\cite{fonseca2022factorizing}. In addition to these, we also incorporate abstracts generated by leading LLMs, including Llama2-70b~\cite{touvron2023llama}, GPT-3.5, and GPT-3.5 with few-shot learning. These models allow us to explore a broad range of summarization techniques, spanning both traditional and cutting-edge approaches.
For each paper in the PubMed dataset, we utilize pretrained models BigBird-PEGASUS-large~\cite{zaheer2020big} and LongT5-large~\cite{guo2022longt5}, as recommended by~\cite{krishna2023longeval}. 
Interestingly, LongT5 and BigBird exhibit a higher degree of extractiveness than human-generated PubMed summaries by default. 
Specifically, they demonstrate 87\% and 74\% bigram overlap with the source text, respectively, compared to the 54\% overlap seen in human-written summaries. 
\textcolor{black}{To introduce diversity in the evaluated summaries, we incorporate the "block" versions of LongT5 and BigBird, which restrict the direct copying of 6-grams from the source text. This approach encourages more abstractive summaries, allowing for a broader evaluation that covers both extractive and abstractive summarization styles.}
Additionally, we experimented with structure-based summarization models, such as those proposed by~\citet{xu2019scientific}, but found their performance to be suboptimal, likely due to the lack of a pretrained backbone in their modified attention mechanism. This limitation impeded their ability to handle complex scientific texts effectively.
In total, our benchmark comprises 500 abstracts generated by various summarization systems, providing a comprehensive evaluation of different methodologies across both the arXiv and PubMed datasets.

\textcolor{black}{\subsection{Human Evaluation Process}} 

\begin{table}[ht]
  \centering
  
  \caption{Inter-annotator agreement between experts (Cohen's $\kappa$ and proportion of agreement).}
  \begin{tabular}{cccc}
    \toprule
    Facet & Classes & Agreement & $\kappa$ \\
    \midrule
    Background & 3 & 0.91 & 0.83 \\
    Method & 4 & 0.78 & 0.69 \\
    Result & 4 & 0.86 & 0.79 \\
    Conclusion & 3 & 0.90 & 0.85 \\
    \bottomrule
  \end{tabular}
  \label{tab:iaa}
\end{table}

\label{annotation}
We have two annotators, who are PhDs with expertise in both bioinformatics and computer science. Together, they select cases to serve as in-context examples for a few-shot learning setting. Subsequently, they independently annotated all cases, evaluating pairs of (target, generated) summaries from a paper based on four facets. Whenever there were differences in their scores, the annotators engaged in discussion to reach a consensus. This annotation process also aligns with previous studies~\cite{jin2019pubmedqa,Liu2022RevisitingTG}.

We also calculate the initial inter-annotator agreement before discussion, measured by Cohen's Kappa and agreement proportions for all four facets, as shown in Table \ref{tab:iaa}. Overall, the agreement rates exceeded those in previous datasets like ACU and medical literature reviews~\cite{wang2023automated}. Notably, the method and result facets showed lower agreement, consistent with the expectation of their varied classification levels.

To assess the relative importance of different facets in the abstract and compute an overall score, we conducted an additional annotation step where one annotator assigned overall scores to each summary. These overall scores reflect the general alignment and quality of the generated abstract compared to the target reference, incorporating aspects such as coherence, correctness, and completeness.
We derived the weights for each facet through a linear regression process, where the overall scores served as the dependent variable and the individual facet scores (background, method, result, conclusion) acted as predictors. The resulting weights of $[0.1, 0.3, 0.3, 0.3]$ indicate the contribution of each facet to the overall score. The low mean squared error of 0.005 demonstrates the effectiveness of the linear fit, confirming that the weights reflect meaningful differences in how facets influence the overall evaluation.

\section{Benchmark Analysis}

\label{sec:experiment}

\begin{table*}[t!]
\centering
\addtolength{\tabcolsep}{-1pt}
\caption{Performance of various summarization systems in different metrics on PubMed dataset. \colorbox{blue!10}{black} cells indicate the best result, while \colorbox{yellow!25}{yellow} cells denote the second best. In general, the smaller LongT5 competes well with Llama2 across different metrics. Specifically, FM-based methods and human annotation tend to favor Llama2, in contrast to existing metrics that primarily rely on $n$-gram overlap calculations similar to ROUGE.}
\resizebox{\textwidth}{!}{%
\begin{tabular}{lccccccccc}
\toprule
 & \multicolumn{2}{c}{Single-score} & \multicolumn{2}{c}{QA/Verification-based} & \multicolumn{1}{c}{} & \multicolumn{3}{c}{Facet-aware} & \\ 
Model & ROUGE-L & BERTScore & ACU & QuestEval & G-EVAL & FM(Llama2) & FM(GPT-3.5) & FM(GPT-4) & Human \\
\midrule
GPT-3.5 & 0.2109 & \cellcolor{yellow!25}0.8408 & 0.1914 & 0.2333 & \cellcolor{blue!10}0.9143 & 0.7691 & 0.6343 & 0.6623 & 0.6780 \\
Llama2 & 0.2223 & \cellcolor{yellow!25}0.8408 & 0.2126 & \cellcolor{yellow!25}0.2678 & 0.7633 & \cellcolor{blue!10}0.8769 & \cellcolor{blue!10}0.7228 & \cellcolor{blue!10}0.7120 & \cellcolor{blue!10}0.7704 \\
LongT5 & \cellcolor{blue!10}0.2832 & \cellcolor{blue!10}0.8534 & \cellcolor{blue!10}0.2533 & \cellcolor{blue!10}0.2699 & \cellcolor{yellow!25}0.8367 & \cellcolor{yellow!25}0.7719 & \cellcolor{yellow!25}0.6591 & \cellcolor{yellow!25}0.6818 & \cellcolor{yellow!25}0.7241 \\
LongT5-block & \cellcolor{yellow!25}0.2345 & \cellcolor{yellow!25}0.8408 & \cellcolor{yellow!25}0.2128 & 0.2496 & 0.4939 & 0.7207 & 0.6283 & 0.6628 & 0.6782 \\
BigBird & 0.2240 & 0.8317 & 0.2127 & 0.2376 & 0.4939 & 0.6687 & 0.5947 & 0.5649 & 0.6186 \\
BigBird-block & 0.2127 & 0.8383 & 0.1918 & 0.2392 & 0.4327 & 0.7347 & 0.6475 & 0.6167 & 0.6317 \\
\bottomrule
\end{tabular}
}
\addtolength{\tabcolsep}{1pt}
\label{tab:systems_updated}
\end{table*}

\subsection{Comparing Summarization Systems}
\label{compare_system}

\begin{table*}[tb]

\centering
\addtolength{\tabcolsep}{-1pt} 
\caption{Performance of various summarization systems in different metrics on arXiv dataset.
\cellcolor{blue!10}Blue indicates the best result, while \cellcolor{yellow!25}yellow denotes the second best. 
Generally, all metrics favor Llama2 and FactSum. 
Specifically, FM-based methods tend to favor Llama2, in contrast to existing metrics that primarily rely on $n$-gram overlap calculations similar to ROUGE.
}
\begin{tabular}{lcccccccc}
\toprule
 & \multicolumn{2}{c}{Single-score} & \multicolumn{3}{c}{QA/Verification-based} & \multicolumn{3}{c}{Facet-aware} \\
Model & ROUGE-L & BERTScore & DELTA & QuestEval & ACU & FM(GPT-3.5) & FM(GPT-4) & Human \\
\midrule
GPT-3.5 & 0.2023 & 0.8337 & 0.2553 & 0.1322 & 0.0552 & 0.6195 & 0.6092 & 0.6385 \\
Llama2 & \cellcolor{yellow!25}0.2338 & 0.8367 & \cellcolor{yellow!25}0.2593 & \cellcolor{yellow!25}0.1742 & 0.0000 & \cellcolor{blue!10}0.6621 & \cellcolor{blue!10}0.6915 & \cellcolor{blue!10}0.7155 \\
FactSum & \cellcolor{blue!10}0.3089 & \cellcolor{blue!10}0.8664 & \cellcolor{blue!10}0.3209 & \cellcolor{blue!10}0.2129 & \cellcolor{blue!10}0.1623 & \cellcolor{yellow!25}0.6536 & \cellcolor{yellow!25}0.6863 & \cellcolor{yellow!25}0.6843 \\
BART-Large & 0.2270 & \cellcolor{yellow!25}0.8495 & 0.2597 & 0.1494 & \cellcolor{yellow!25}0.0907 & 0.5912 & 0.5785 & 0.6231 \\
\bottomrule
\end{tabular}
\addtolength{\tabcolsep}{1pt} 
\label{tab:arxiv_system} 
\end{table*}

In Table ~\ref{tab:systems_updated}, we show the performance of the summarization systems in different metrics on the PubMed dataset.
We do not include GPT-3.5's with few-shot learning, as it does not improve performance.
Similar results on arXiv are in Table ~\ref{tab:arxiv_system}.
Generally, GPT-3.5, Llama2, and Long5 consistently achieve higher scores across all metrics, showing their robustness and adaptability in different domains.
Specifically, Llama2 shows the highest performance, similar to the observation in the news domain~\cite{website}. 
This highlights the potential of applying open-source LLMs as alternatives to closed LLMs.
The result also suggests that \textit{finetuned smaller-scale models can rival the performance of LLMs in scientific contexts.}
However, achieving such performance demands precise model design, as seen in BigBird's inferior performance compared to LLMs.

\begin{figure}[htb]
\centering
\includegraphics[scale=0.61 ]{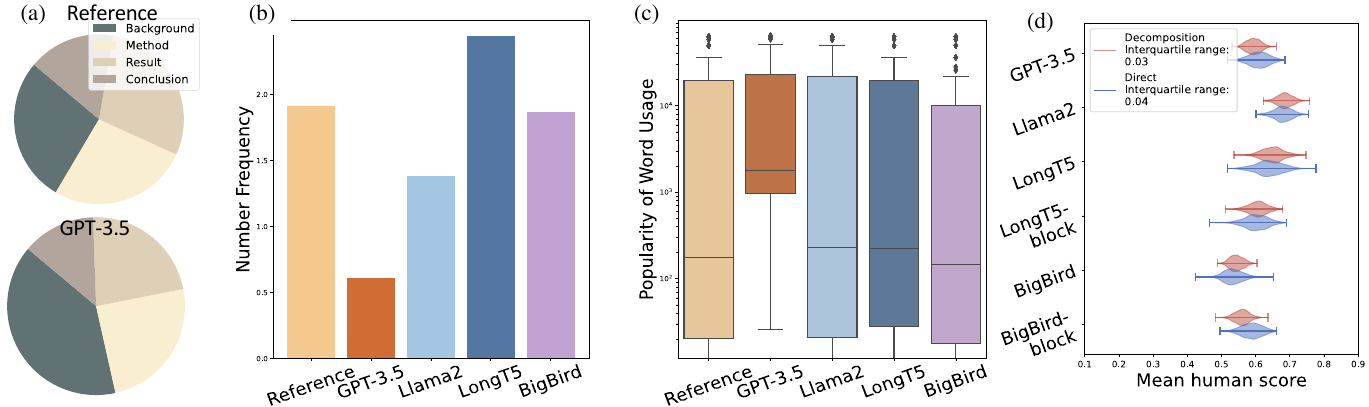}
\caption{
(a) Proportions of different facets. (b) Frequency of numbers in reference and generated text. (c) Popularity of word usage. (d) Mean human evaluation score distributions for various models shown by violin plots, comparing two annotation methods through bootstrap resampling. 
}
\label{fig:histogram}
\end{figure}

We also present a comprehensive statistical evaluation in Figure \ref{fig:histogram}. Our analysis reveals that, \textit{while GPT-3.5 tends to produce text that is generally more accessible and easier to comprehend, it frequently omits critical scientific statistics}. As illustrated in Figure \ref{fig:histogram}(a), the model often generates an extensive background in the abstract, yet it incorporates significantly fewer numerical details compared to other models, as depicted in Figure \ref{fig:histogram}(b). This discrepancy is especially concerning given the pivotal role that quantitative data plays in scientific literature. Moreover, GPT-3.5 demonstrates a preference for more commonly used words, a pattern that becomes apparent in Figure \ref{fig:histogram}(c).
Further human evaluation of different qualitative aspects is detailed in Appendix~\ref{appendix_facet}. Case studies suggest that when conclusions diverge from established background information, GPT-3.5 often adheres to conventional knowledge rather than adapting to align with the newly presented conclusions. Additionally, our statistical analysis reveals that 34.7\% of the model’s weaker performances in pre-trained language models (PLMs) are associated with difficulties in maintaining fluency, particularly when generating longer text segments in final conclusions.

\subsection{Comparison of FM Metric Against Baseline Metrics}

\label{main_metric}

% We next assess the evaluation metrics.

\begin{figure}[ht]
\centering
\includegraphics[width=0.49\textwidth]{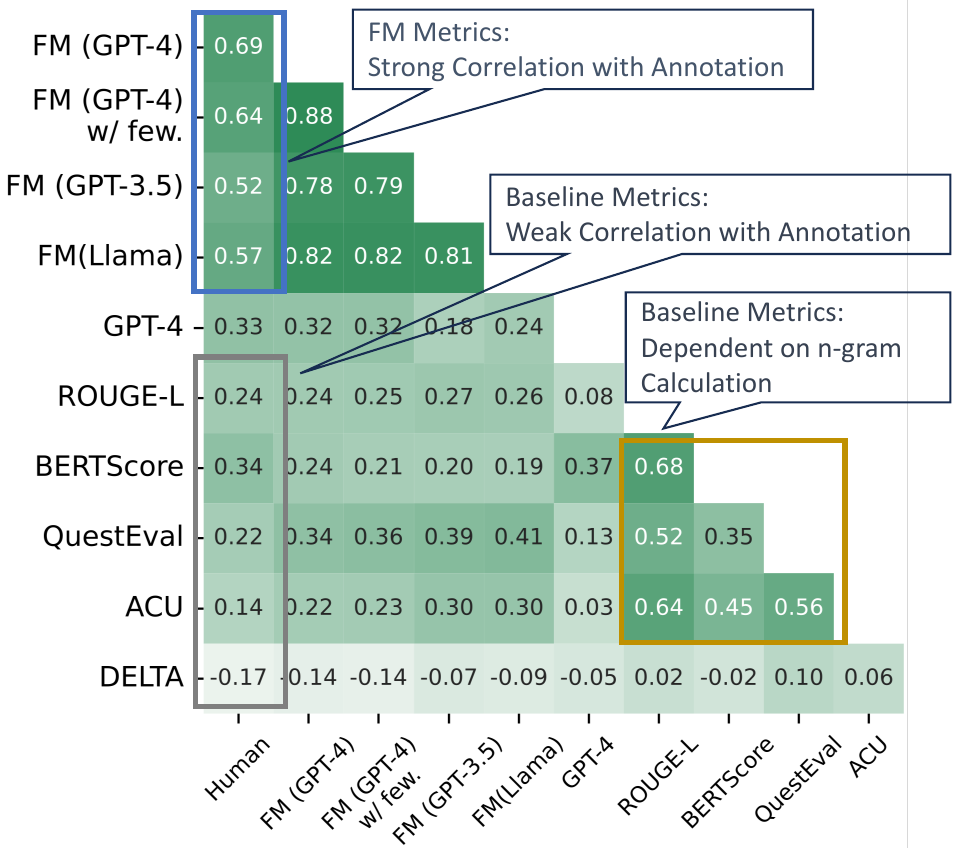}
\caption{
 Spearman correlations among metrics within our FM paradigm, LLM-based baseline (GPT-4), and existing evaluation metrics (ROUGE-L, etc). 
}
\label{fig:metrics}
\end{figure}

\begin{figure}[htb]
\centering
\includegraphics[scale=0.49]{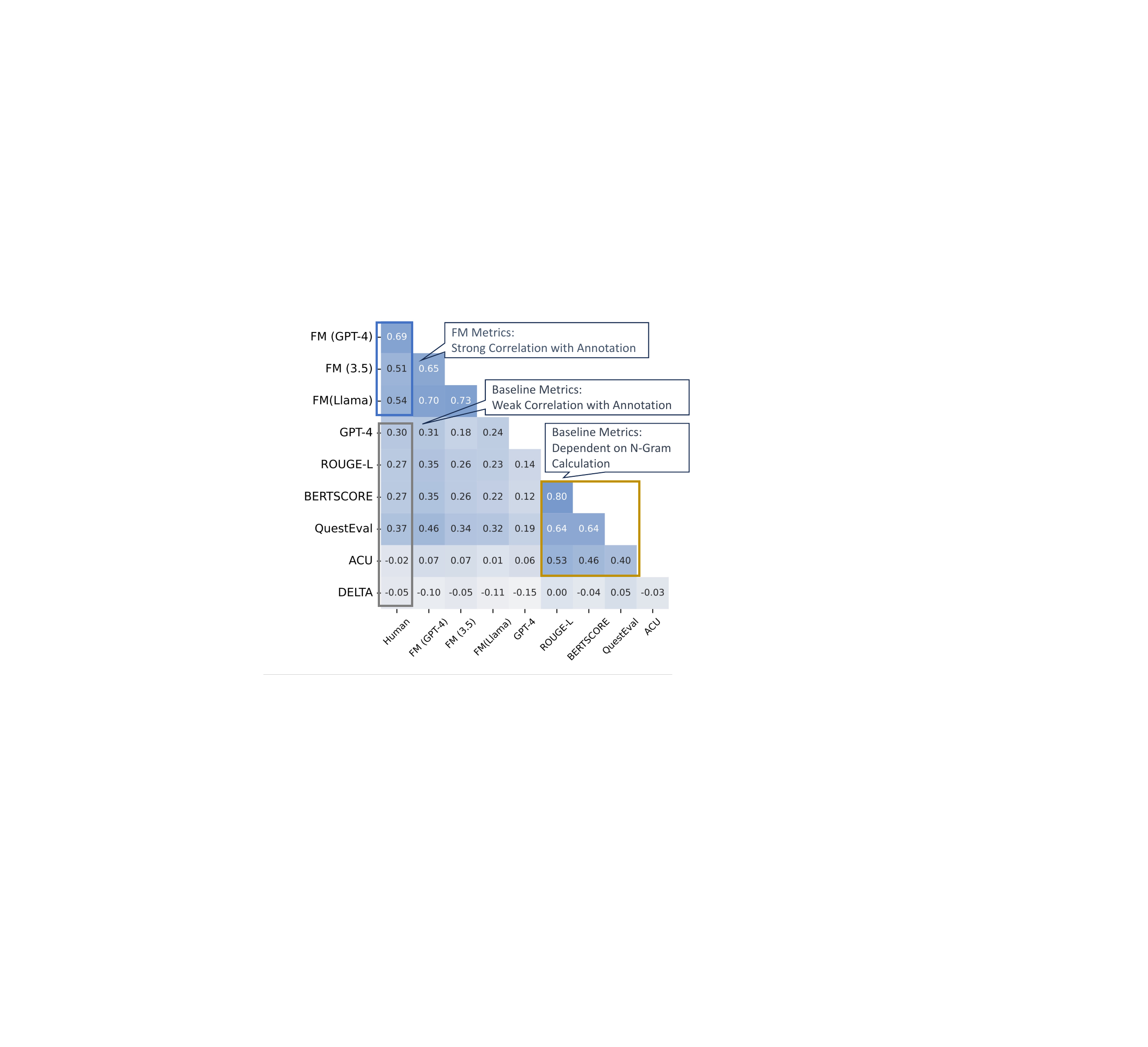}
\caption{
 Spearman correlations among metrics within our FM paradigm, LLM-based baseline (GPT-4), and existing evaluation metrics (ROUGE-L, etc) on arXiv dataset.
}
\label{fig:arxiv_metrics}
\end{figure}

\textbf{Benefits of Our FM.}
We can draw several conclusions based on the evaluation performance.
Firstly, \textit{the existing evaluation metrics show a moderate correlation with human scores}, as seen in the gray box of Figure \ref{fig:metrics}, with correlations below 0.4, and further confirmed by Table \ref{tab:systems_updated}. 
For example, BERTScore's similar ratings for different models reveal its limited differentiation capability. 
G-EVAL also faces challenges, notably favoring GPT-generated text significantly.
Results on arXiv dataset is shown in Figure~\ref{fig:arxiv_metrics}.
TIGERSCORE, not included in the figure, consistently fails to produce scores, highlighting the need for robustness and generalizability in embedding-based metrics, especially in specialized domains.

Secondly, \textit{our decomposed evaluation paradigm simplifies the evaluation task for LLMs without requiring advanced reasoning capabilities}. 
As highlighted in the blue box in Figure \ref{fig:metrics}, our FM family of metrics consistently demonstrates a strong correlation with human evaluations, achieving an impressive correlation of up to 0.69. 
This result holds true not only for powerful closed-source LLMs like GPT-4 but also for open-source models such as LLaMA, showcasing the generalization and applicability of our approach. 
Conversely, when GPT-4 and LLaMA2 are evaluated without applying the decomposition strategy, they fail to achieve similarly high correlations, underscoring the importance of our proposed evaluation framework.

Thirdly, \textit{our approach offers a deeper semantic analysis beyond mere $n$-gram overlaps.} Traditional metrics such as BERTScore, QuestEval, and ACU, while widely used, tend to correlate strongly with ROUGE-L (as shown by the yellow box in Figure \ref{fig:metrics}). These metrics consistently favor LongT5 as the top-performing model (Table \ref{tab:systems_updated}), which suggests that their evaluation is largely based on surface-level word sequence overlaps rather than a comprehensive understanding of the text. 
\textcolor{black}{In contrast, our FM-based metrics rank Llama2 higher, aligning more closely with human judgments}
This demonstrates the ability of our method to reflect a more meaningful understanding of the content beyond simple token matching.

\textbf{LLM Analysis.}
Unlike traditional approaches that assess LLMs through direct question-answering, our framework employs a meta-evaluation method to examine LLMs' evaluative capabilities.
\textcolor{black}{As shown in Figure~\ref{fig:metrics}, our analysis reveals that few-shot learning prompts fail to enhance the performance of both GPT-3.5 and GPT-4 in evaluation tasks. 
Moreover, introducing few-shot learning in the summarization process also fails to yield noticeable improvements. 
These findings suggest that LLMs have inherent limitations in acquiring and applying scientific knowledge through in-context learning, particularly when transferring few-shot learning from one scholarly issue to another.}
This highlights a critical area for future enhancement, as addressing these limitations could significantly improve the applicability of LLMs in scientific domains where precise knowledge understanding and contextual reasoning are essential.

\textbf{Comparison of QA and Verification-based Methods.} 
When examining verification-based metrics (e.g., ACU) in contrast to question-answering (QA)-based approaches (e.g., QuestEval), it becomes evident that the QA paradigm consistently outperforms its verification-based counterpart across multiple datasets, as illustrated in Figure \ref{fig:metrics}. Notably, QuestEval yields superior performance on both datasets by a significant margin. This outcome underscores the inherent difficulty faced by language models when tasked with breaking down and verifying semantic meaning at a granular level. In contrast, answering specific, concise questions—requiring brief, targeted responses in the form of phrases or keywords—proves to be a more manageable and effective task for these models. Consequently, QA-based approaches demonstrate greater robustness in capturing the nuances of summarization quality. This suggests that \textit{breaking semantic meanings into units is challenging for language models, whereas answering concise questions with brief phrases and words is more straightforward and effective.}

\begin{figure}[htb]
\centering
\includegraphics[scale=0.45]{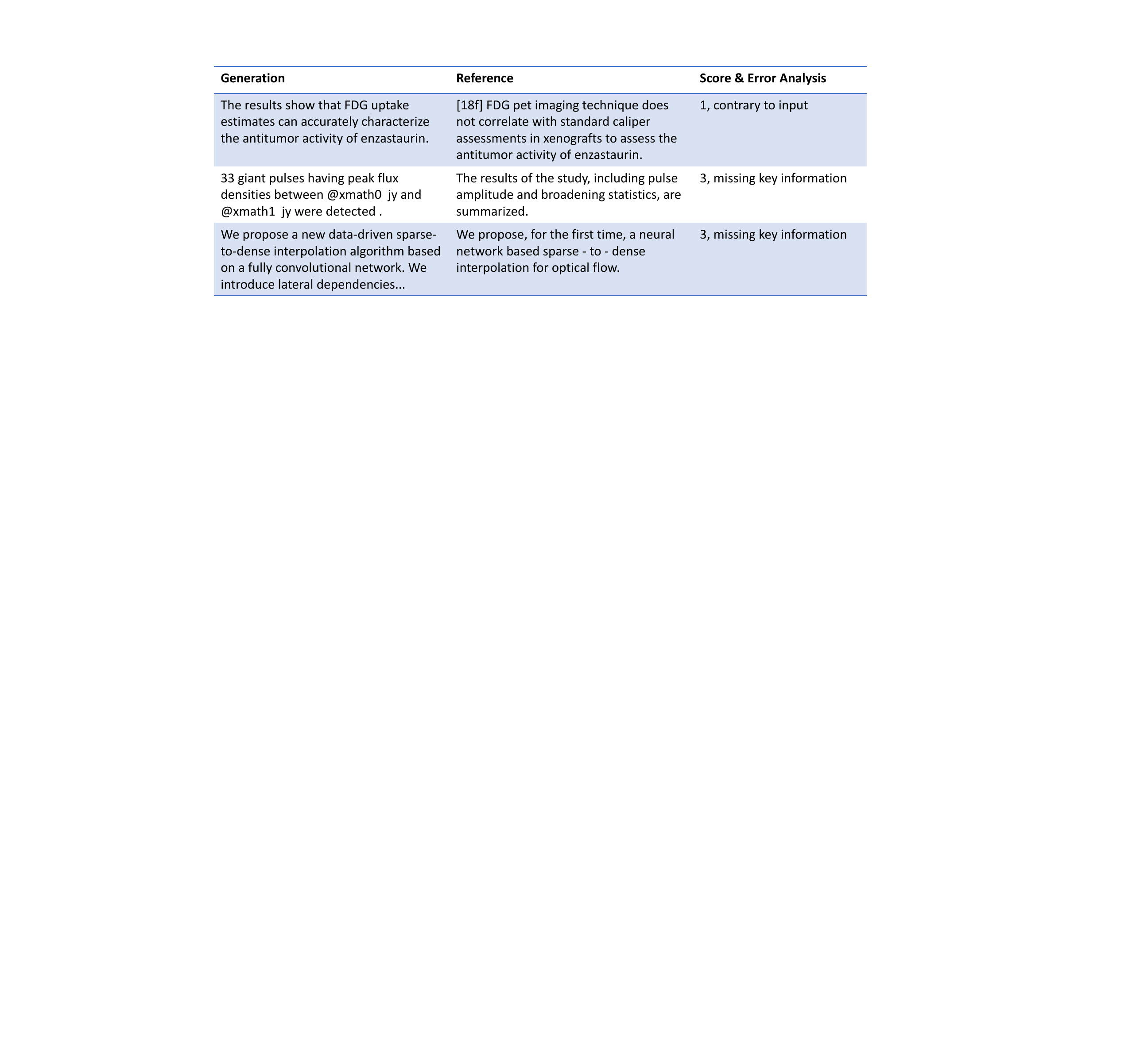}
\caption{
Case study across two datasets of our FM (GPT-4).
}
\label{fig:error}
\end{figure}

\textcolor{black}{\textbf{Case Study.}}
The case study presented in Figure \ref{fig:error} demonstrates several examples where the model’s generated content deviates from the reference text, as evaluated using the FM (GPT-4) metric. 
The first example showcases a case where the generated output incorrectly claims that "FDG uptake estimates can accurately characterize the antitumor activity of enzastaurin." However, the reference explicitly contradicts this statement, noting that FDG PET imaging does not correlate with standard caliper assessments used to evaluate enzastaurin's antitumor activity.
This type of error, classified as "Contrary," represents a significant discrepancy between the generation and the reference, earning the lowest score of 1, indicating a critical misalignment.
The second and third examples highlight instances where the generated content fails to include key information. In one case, the generation mentions "33 giant pulses having peak flux densities," but omits important details about pulse amplitude and broadening statistics that are present in the reference. Similarly, in another example, the generated text introduces a new "data-driven sparse-to-dense interpolation algorithm," yet fails to mention the neural network-based interpolation specifically outlined in the reference. Both of these cases are categorized under "missing key information," receiving a score of 3. While the generated summaries capture some aspects of the reference content, they lack critical details, leading to incomplete representations.

These cases underscore the FM metric’s ability to detect different types of errors, such as generating contradictory information or omitting essential details. By identifying these nuanced mistakes, the FM metric provides a more comprehensive evaluation of model performance compared to traditional metrics, which often overlook such specific shortcomings.

\textbf{Limited Inference Cost.} 
To demonstrate that our FM metric does not significantly increase inference costs, we compare it with the LLM-based G-Eval. G-Eval uses separate evaluation prompts for relevance, fluency, coherence, and consistency, which requires inputting the document four times. This repeated evaluation results in a nearly 5x increase in computational overhead, leading to much higher costs. In contrast, our FM metric streamlines the process by performing multi-facet evaluation simultaneously through summarization, eliminating the need to repeatedly process the document for each criterion. Instead, it adds only a small number of additional summarization and evaluation prompts. As a result, the overall token count increases by just 10\%, demonstrating a significant improvement in efficiency without sacrificing the depth of evaluation.

\begin{figure}[htb]
\centering
\includegraphics[scale=0.23]{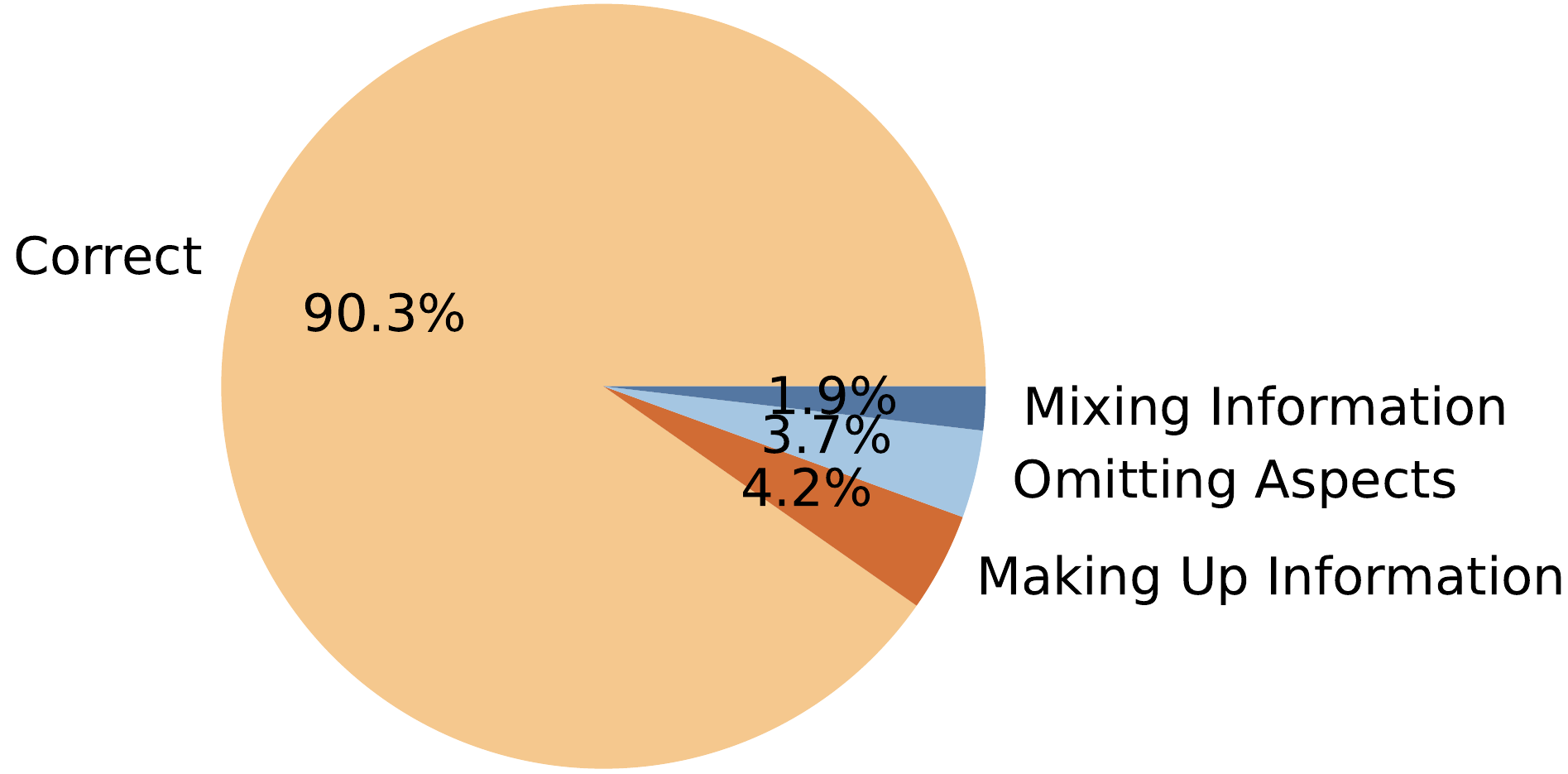}
\caption{
Human evaluation of GPT-4's facet information extraction.
}
\label{fig:pie}
\end{figure}

\textcolor{black}{\subsection{Error Analysis}}
\label{appendix_first}

In this subsection, we analyze the errors encountered by our FM metric, which can occasionally yield incorrect assessments. These errors typically arise in two main scenarios.

The first type of error is related to facet information extraction. 
We conducted a comprehensive human evaluation to assess GPT-4's performance in this task, as illustrated in Figure \ref{fig:pie}. Overall, GPT-4 demonstrates strong performance, achieving an impressive accuracy rate of 90\%. However, certain errors were identified during the evaluation process, including cases where GPT-4 confused or mixed different aspects, omitted critical information, or fabricated details that were not present in the input.

A common issue involves the omission of important aspects from the extracted content. For example, in a paper's conclusion, the original text highlights two key points: (1) significant associations were found between low serum concentrations of zinc and 25(OH)D, and (2) food fortification or mineral supplementation should be considered in future health programs. However, the extracted facet only includes the first point, omitting the second. This omission compromises the comprehensiveness of the evaluation and misrepresents the full scope of the conclusion.

This phenomenon of generating incorrect or non-existent information, commonly known as hallucination, is a well-documented issue in LLMs. While GPT-4’s extraction capabilities are generally robust, these challenges underscore the need for further refinement. We are optimistic that future iterations of LLMs will address these issues more effectively, reducing errors like hallucination and improving overall accuracy and reliability in content extraction tasks. Such advancements are crucial to enhancing the model's usability in real-world applications that require precise and dependable information retrieval.

The second type of error occurs during the scoring process, often due to the evaluation LLM’s limited domain-specific knowledge.
For instance, the ground truth conclusion states: “If the early trends in MHS are indicative of the final results, CMS will face the decision of whether to abandon commercial DM in favor of other chronic care management strategies.” The generated conclusion, on the other hand, states: “The model struggles with a mismatch between short-term ROI expectations and long-term cost-saving goals. Aligning savings timelines with stakeholder needs and targeting cost categories with reduction potential can help policymakers justify and improve the commercial DM model.” Both conclusions share key semantic similarities, but they include several domain-specific abbreviations, such as MHS, CMS, DM, and ROI. The evaluation LLM, due to its limited understanding of these terms, assigned a medium score, failing to fully capture the semantic equivalence between the two conclusions.

We suggest that strategies like domain-specific fine-tuning or the integration of external knowledge could potentially improve the robustness and accuracy of our evaluation paradigm. 

\textcolor{black}{\subsection{Statistical Power Analysis of the Measure}}

\begin{figure}[htb]
\centering
\includegraphics[scale=0.4]{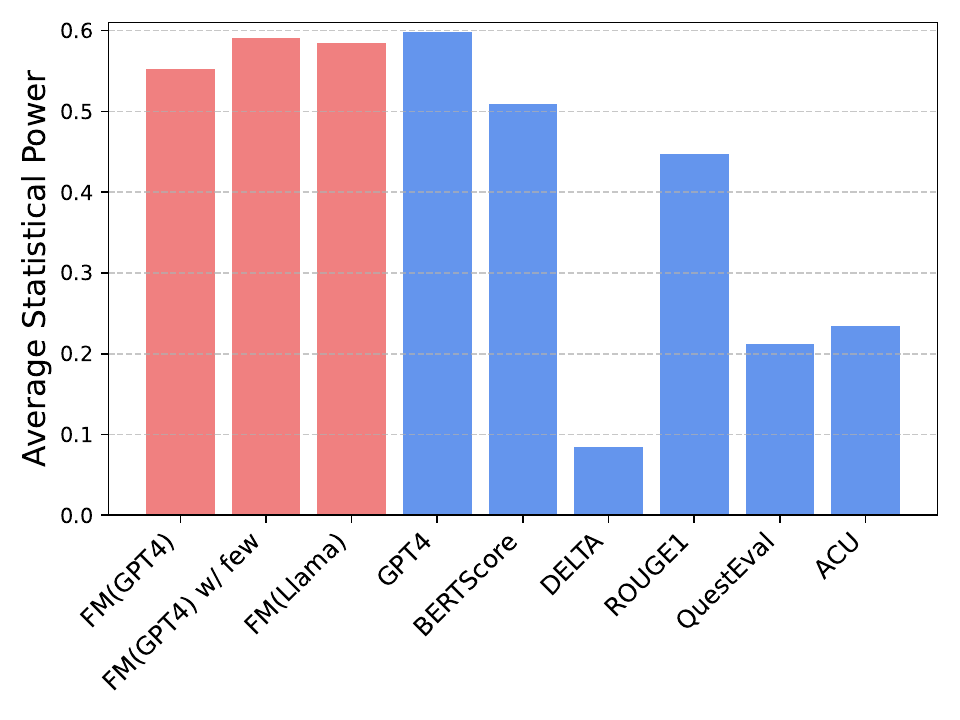}
\caption{
\textcolor{black}{Average statistical power of different metrics, with FM(GPT4) and FM(GPT4) w/ few-shot showing higher power compared to task-specific baselines like QuestEval and ACU.}
}
\label{fig:power}
\end{figure}

We conducted a statistical power analysis to evaluate the reliability and effectiveness of our proposed metrics in detecting meaningful differences between summarization systems, shown in Figure \ref{fig:power}. Using 1,000 bootstrap resampling tests and t-tests ($\alpha$ = 0.05), we compared our framework-enhanced metrics with baseline metrics across multiple biomedical summarization models.

The results indicate that FM(GPT4) and FM(GPT4) with few-shot prompting consistently exhibit higher statistical power compared to traditional task-specific metrics, demonstrating their robustness in detecting system-level differences. In contrast, task-specific metrics like QuestEval and ACU show lower statistical power, limiting their effectiveness in identifying significant differences. Traditional metrics such as ROUGE-1 and BERTScore display moderate sensitivity, while DELTA struggles to capture meaningful differences.

Additionally, we observed consistent performance across the PubMed and arXiv datasets, indicating the generalizability of our proposed metrics. This analysis highlights the importance of proper statistical testing in evaluation frameworks, ensuring reliable and robust performance across diverse data domains.

\subsection{Performance in Different Facets}
\label{appendix_facet}

\begin{figure}[htb]
\centering
\includegraphics[scale=0.4]{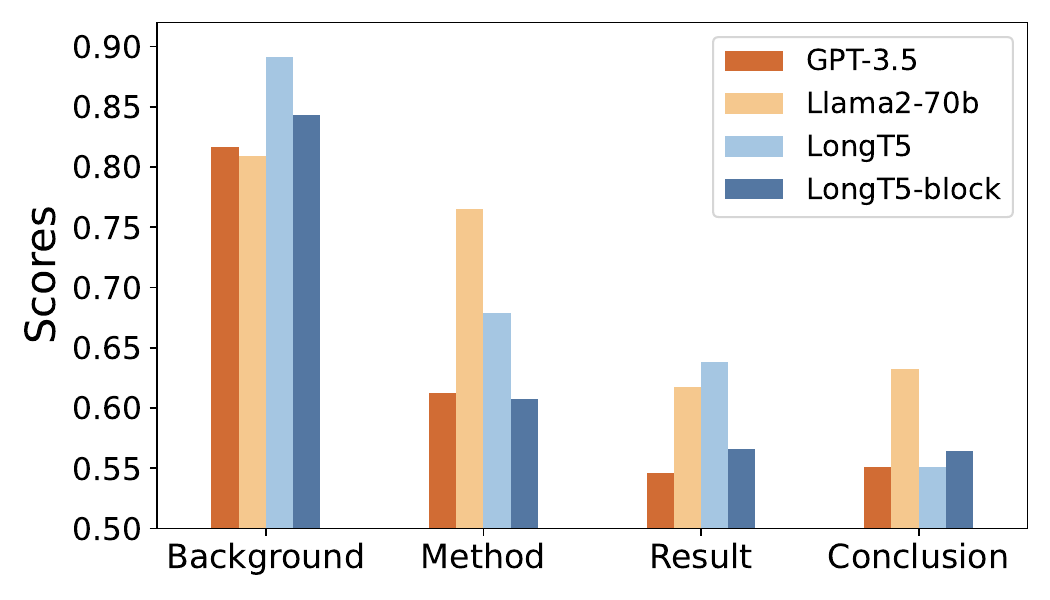}
\caption{
Model performance across four facets.
}
\label{fig:facets}
\end{figure}

Beyond the overall evaluation, we show the human evaluation of the models' performance in various aspects of abstract writing in Figure ~\ref{fig:facets}.
Firstly, all models show higher performance in the background aspect, as it often involves just a broad, less detailed overview.
In contrast, other aspects demand more precise alignment with the input, leading to generally lower model performance in these areas.
Among these three aspects, Llama2 consistently exhibits relatively higher performance.
In comparison, GPT-3.5 and other PLMs exhibit weaker performance, especially in formulating conclusions. 
Specifically, in scenarios where the work's conclusion deviates from conventional results mentioned in the background, \textit{GPT-3.5 can adhere to the conventions instead of being faithful to the conclusion in the input.}
This could be because GPT-3.5 relies more on its internal knowledge base, without thoroughly analyzing the input content.
We additionally have a statistical analysis that reveals 34.7\% of weak performance cases (where the conclusion score is below 3) in PLM models are due to fluency issues in longer text generation in the last conclusion part.

\section{Effectiveness of Granularity-level Decomposition}
\label{sec:discussion}
\subsection{Decomposition in Summary Evaluation}
In Figure \ref{fig:metrics} we show that GPT-4 underperforms our FM (GPT-4) by a large margin, demonstrating the effectiveness of decomposition in automatic evaluation process.
Furthermore, in Figure \ref{fig:histogram}(d), we present a violin plot comparing human evaluation scores using decomposition and direct annotation methods. The plot compares two annotation paradigms: 1) allowing humans to directly assign an overall score to each summary, and 2) having humans evaluate each aspect based on our facet-aware paradigm.
These plots are generated through bootstrap resampling, a robust method for assessing score consistency over multiple evaluations~\cite{krishna2023longeval,cohan2016revisiting}. 
\textit{The decomposition method demonstrates a significant advantage in producing reliable and consistent annotations}, as evidenced by its considerably narrower interquartile range (0.26 compared to 0.40). 

It is also crucial to highlight that, despite the methodological differences between the decomposition approach and direct annotation methods, both techniques consistently produce the same relative ordering of systems. This consistency strongly indicates that the model scoring is not influenced by the particular method of evaluation employed; rather, it remains stable and reliable across various annotation strategies. This observation further emphasizes the robustness of our approach, as it demonstrates that the metrics aligned with our decomposition paradigm are not only consistent with human evaluation results conducted within the same framework but also show a strong correlation with the gold standard evaluation. Such alignment reinforces the credibility of our evaluation process and validates that our metrics are both reliable and in line with well-established evaluation benchmarks, further showcasing the overall effectiveness and adaptability of our paradigm in assessing system performance.
\begin{figure}[htb]
\centering
\includegraphics[scale=0.4]{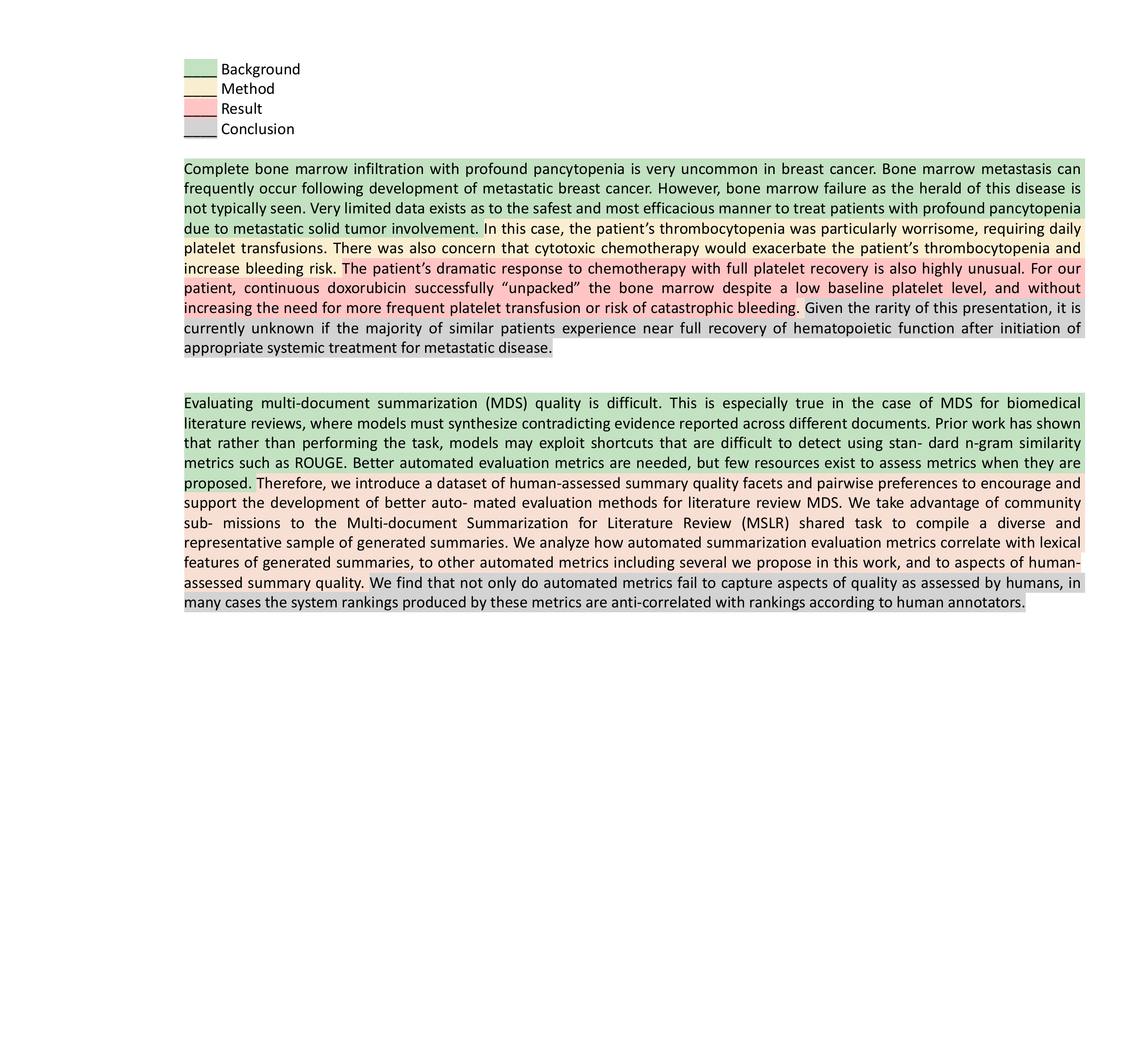}
\caption{
Highlight visualization for reading summaries during the question-answering task.
}
\label{fig:rate}
\end{figure}

\textbf{Decomposition with non-LLM metrics.} 
We broadened the scope of our decomposition paradigm by integrating non-LLM metrics into the evaluation process, creating a more comprehensive and nuanced assessment framework. Specifically, we utilized GPT-4 to extract and distill content related to key aspects of our data, such as factual accuracy, coherence, and relevance. These extracted elements were then evaluated using established, non-LLM metrics like ROUGE and BERTScore, allowing for a dual-layered evaluation that combines the interpretative power of large language models with the objectivity of traditional performance metrics.
This hybrid approach facilitated a more granular and objective assessment of various components, leading to insightful comparisons between model outputs. The results were compelling: we observed a significant 8.3\% improvement in ROUGE scores, which increased from 0.24 to 0.26. Additionally, the Spearman correlation, a metric that assesses the rank correlation between predicted and ground truth values, saw an 8.8\% enhancement, with average values rising from 0.34 to 0.37.
These improvements highlight the effectiveness of blending LLM-driven content extraction with time-tested, non-LLM evaluation methods. By leveraging the complementary strengths of both paradigms, our framework not only improves overall performance but also enhances the robustness and reliability of the evaluation process. This demonstrates that a hybrid model, combining advanced LLM capabilities with traditional metrics, offers a more balanced and effective approach to performance assessment, enabling a deeper understanding of the models' strengths and weaknesses across different dimensions.

\subsection{Decomposition in Summary Reading}

Apart from the decomposition in the evaluation process, we are also interested to see the role of decomposition in user reading process, to assess the potential benefits of this method.

Concretely, we conducted a user study involving six PhD participants who were tasked with reading papers sampled from two datasets. Each participant skimmed the abstract of a paper and answered two multiple-choice questions related to its content. We tracked both the time taken to respond and the accuracy of their first responses, aiming to assess the impact of our decomposition markers on reading efficiency and comprehension. Details of the rating interface used in this study can be found in Figure \ref{fig:rate}.

The results showed that participants using our decomposition markers were able to answer the questions significantly faster (average time $\mu$ = 47.9s, standard deviation $\sigma$ = 20.8s) compared to those using a standard document reader ($\mu$ = 55.0s, $\sigma$ = 23.2s). This difference in speed was statistically significant, with a p-value of less than 0.05. Moreover, 4 out of 6 annotators reported that the decomposition process made the task easier and more manageable.
Importantly, this improvement in time efficiency did not come at the cost of accuracy. The accuracy rates for participants using decomposition markers ($\mu$ = 0.82, $\sigma$ = 0.34) were not significantly different from those for participants reading plain text ($\mu$ = 0.79, $\sigma$ = 0.31). 
This finding suggests that while decomposition improves reading speed, it does not negatively affect the reader's ability to correctly comprehend the content. Therefore, the results demonstrate that \textit{decomposition on abstract reading is also beneficial for the reading process}~\cite{sollaci2004introduction}, providing both efficiency and ease of understanding without compromising accuracy.

This study highlights the potential of our approach to enhance reading and comprehension tasks, particularly for complex academic papers, where rapid understanding is often crucial.

\section{ScholarSum dataset release}
\label{release}
\subsection{Accessing ScholarSum}
The dataset source files are stored in JSON format, and they are uploaded to GitHub and can be downloaded publicly. 
A case in MDS consists of article, summary, and annotations. 
Below is a high-level structure:
\begin{tcolorbox}[colback=gray!10, left=1mm, right=1mm, top=1mm, bottom=1mm,breakable] 
\small
\textbf{Original Content}\\
- \textbf{human}: Human-written summary of the article.\\
- \textbf{article}: The original article content.\\

\textbf{Model-Generated Summaries}\\
- \textbf{bartbase}: Summary generated by the BARTbase model.\\
- \textbf{bartlarge}: Summary generated by the BARTlarge model.\\
- \textbf{llama2\_70b}: Summary generated by the Llama2\_70b model.\\
- \textbf{gpt35}: Summary generated by the GPT-3.5 model.\\
- \textbf{factsum}: Summary generated by the FactSum model.\\

\textbf{Model-Generated Aspects}\\
- \textbf{human\_aspect}: Aspects of the article as understood and annotated by humans.\\
- \textbf{gpt35\_aspect}: Aspects of summary generated by the GPT-3.5 model.\\
- \textbf{llama2\_70b\_aspect}: Aspects of summary generated by the Llama2\_70b model.\\
- \textbf{bartbase\_aspect}: Aspects of summary generated by the BARTbase model.\\
- \textbf{bartlarge\_aspect}: Aspects of summary generated by the BARTlarge model.\\
- \textbf{factsum\_aspect}: Aspects of summary generated by the FactSum model.\\

\textbf{Human and Model Evaluation Scores}\\
- \textbf{gpt35\_human}: Evaluation score given by humans for the GPT-3.5 model.\\
- \textbf{gpt35\_human\_list}: A list of human evaluation scores for different aspects of the GPT-3.5 model.\\
- \textbf{gpt35\_gpt4\_fm}: Evaluation score given by GPT-4 for the GPT-3.5 model.\\
- \textbf{gpt35\_gpt4\_fm\_list}: A list of GPT-4 evaluation scores for different aspects of the GPT-3.5 model.\\
- (Other similar evaluation scores for different models)
\end{tcolorbox}

All annotations, code, and datasets are available at \url{https://github.com/iriscxy/ScholarSum}.

\subsection{ScholarSum Distribution and Maintenance}

\textbf{License}. ScholarSum is distributed under the Creative Commons (CC) copyright licenses.
It is important to note that the source documents used in the dataset are already in the public domain, thereby respecting copyright regulations. 
To address any concerns related to personal or copyrighted content within ScholarSum, we have implemented a contact form on our website. 
This form serves as a channel for users to submit requests for the removal or blacklisting of specific links or content that may infringe upon personal rights or copyrights. 
We are committed to promptly and diligently processing these requests to maintain the integrity and legality of the dataset. 
The authors bear all responsibility in case of violation of rights and confirm the dataset licenses.

\textbf{Maintenance}. The authors are committed to providing long-term support for the ScholarSum dataset. At present, ScholarSum files are hosted on GitHub, allowing for easy access and collaboration. 
Additionally, the authors are committed to actively monitoring the usage of the dataset and addressing any issues that may arise. 
This includes promptly addressing bug fixes, resolving technical concerns, and providing necessary updates to ensure the dataset remains reliable and useful to the research community.

\section{Conclusion \& Limitation}
\label{sec:conclusion}
In this study, we analyze the shortcomings of current summarization evaluation metrics in academic texts,
particularly in providing explanations, grasping scientific concepts, or identifying key content.
We then propose an automatic, decomposable, and explainable evaluation metric, leveraging LLMs for semantic matching assessments. 
We also introduce the first benchmark dataset spanning two scholarly domains. 
Our study highlights significant gaps between automated metrics and human judgment, with our metric aligning more closely with the ground truth.
We also uncovered numerous insightful findings for summarization and evaluation of scholar papers.
We hope our benchmark inspires better evaluation metrics and future enhancements to LLMs.
% Looking ahead, a limitation of our current approach is that it does not cover multi-reference or reference-free evaluation techniques. 
% Our future work aims to explore these methods in the scientific field.
% More discussion can be found in the Appendix.

From limitation aspect, our evaluation metrics rely on the presence of reference summaries, primarily due to the existence of accurate and faithful abstracts for scientific papers.
Nonetheless, our ultimate goal is to assess summary quality without the need for references. 
There are existing reference-free summarization evaluation techniques, but the performance of these metrics in scientific summarization evaluation has yet to be studied, marking an area for future research.
Meanwhile, it's worth noting that a single paper could have several fitting abstracts. 
While our evaluation criteria take into account the varied ways one might craft a competent abstract, having a broader set of human-composed abstracts as a benchmark would be advantageous. 
Our approach is flexible enough to work with multiple references, and we plan to explore frequency modulation using various sources in our future research.

\section*{Acknowledgments}
Xiuying Chen was supported by Mohamed bin Zayed University of Artificial Intelligence (MBZUAI) through grant award 8481000078.
% The work was supported by King Abdullah University of Science and Technology (KAUST) through grant awards FCC/1/1976-44-01, FCC/1/1976-45-01, REI/1/5234-01-01, and RGC/3/4816-01-01.

\clearpage

\bibliographystyle{ACM-Reference-Format}
\bibliography{stickerPreference}

\end{document}